\documentclass[preprint]{vgtc}               

\ifpdf
  \pdfoutput=1\relax                   
  \usepackage{graphicx}                
  \DeclareGraphicsExtensions{.pdf,.png,.jpg,.jpeg} 
\else
  \usepackage{graphicx}                
  \DeclareGraphicsExtensions{.eps}     
\fi%

\graphicspath{{figures/}{pictures/}{images/}{./}} 

\usepackage{microtype}                 
\PassOptionsToPackage{warn}{textcomp}  
\usepackage{textcomp}                  
\usepackage{mathptmx}                  
\usepackage{times}                     
\usepackage{cite}                      
\usepackage{tabu}                      
\usepackage{booktabs}                  

\onlineid{1006}

\vgtccategory{Research}

\vgtcinsertpkg

\ieeedoi{10.1109/VIS54172.2023.00058}

\usepackage{listings}
\usepackage{xcolor}
\usepackage{amsmath}
\usepackage{cleveref}

\usepackage{amsmath}

\colorlet{punct}{red!60!black}
\definecolor{background}{HTML}{EEEEEE}
\definecolor{delim}{RGB}{20,105,176}
\colorlet{numb}{magenta!60!black}

\lstdefinelanguage{json}{
    basicstyle=\normalfont\ttfamily,
    numbers=left,
    numberstyle=\scriptsize,
    stepnumber=1,
    numbersep=8pt,
    showstringspaces=false,
    breaklines=true,
    frame=lines,
    backgroundcolor=\color{background},
    literate=
     *{0}{{{\color{numb}0}}}{1}
      {1}{{{\color{numb}1}}}{1}
      {2}{{{\color{numb}2}}}{1}
      {3}{{{\color{numb}3}}}{1}
      {4}{{{\color{numb}4}}}{1}
      {5}{{{\color{numb}5}}}{1}
      {6}{{{\color{numb}6}}}{1}
      {7}{{{\color{numb}7}}}{1}
      {8}{{{\color{numb}8}}}{1}
      {9}{{{\color{numb}9}}}{1}
      {:}{{{\color{punct}{:}}}}{1}
      {,}{{{\color{punct}{,}}}}{1}
      {\{}{{{\color{delim}{\{}}}}{1}
      {\}}{{{\color{delim}{\}}}}}{1}
      {[}{{{\color{delim}{[}}}}{1}
      {]}{{{\color{delim}{]}}}}{1},
}

\title{ScatterUQ: Interactive Uncertainty Visualizations for Multiclass Deep Learning Problems}

\author{
Harry X. Li\thanks{e-mails: harry.li, steven.jorgensen, john.holodnak, allan.wollaber @ll.mit.edu}
\and Steven Jorgensen
\and John Holodnak
\and Allan B. Wollaber
}
\affiliation{\scriptsize MIT Lincoln Laboratory}

\teaser{
  \centering
  \includegraphics[clip, trim=0cm 0cm 0cm 0cm, width=\linewidth]{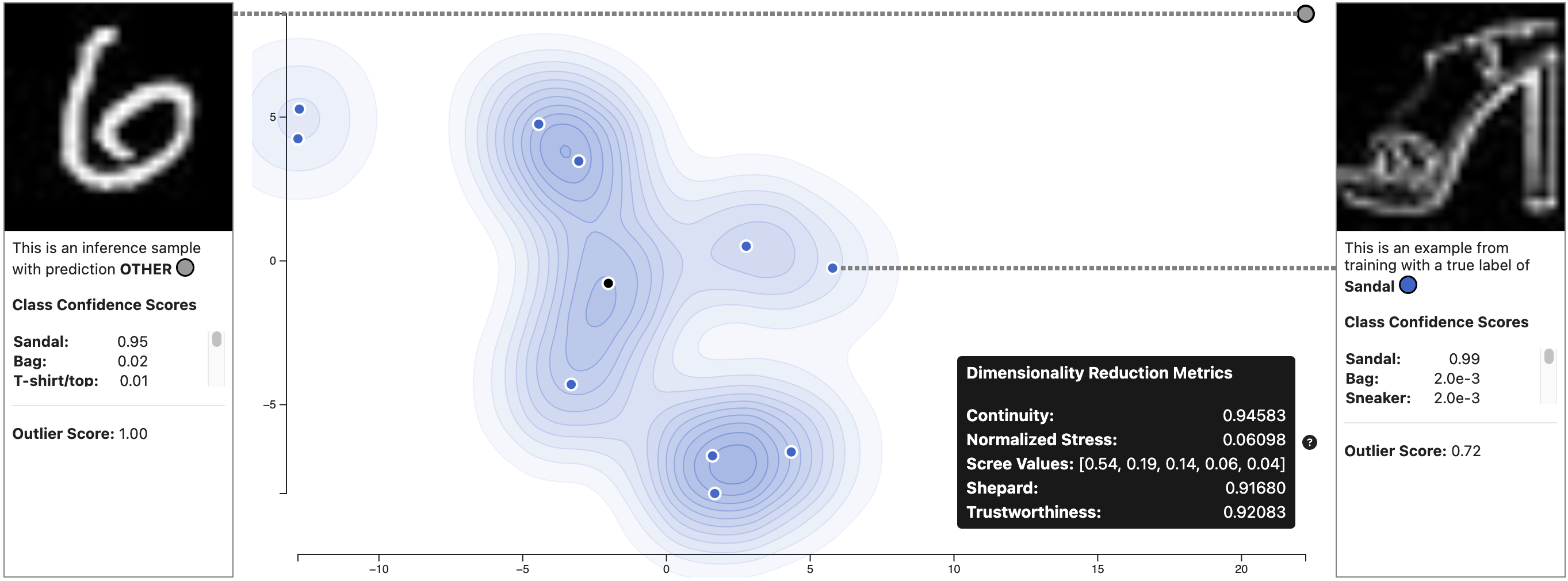} 
  \caption{ScatterUQ plot of an out of distribution MNIST test sample (left sidebar and gray dot), ten Fashion-MNIST training examples from the closest class Sandal (blue dots), and the nearest Sandal training example (right sidebar).}
  \label{fig:ood}
}

\abstract{Recently, uncertainty-aware deep learning methods for multiclass labeling problems have been developed that provide calibrated class prediction probabilities and out-of-distribution (OOD) indicators, letting machine learning (ML) consumers and engineers gauge a model's confidence in its predictions. However, this extra neural network prediction information is challenging to scalably convey visually for arbitrary data sources under multiple uncertainty contexts. To address these challenges, we present ScatterUQ, an interactive system that provides targeted visualizations to allow users to better understand model performance in context-driven uncertainty settings. ScatterUQ leverages recent advances in distance-aware neural networks, together with dimensionality reduction techniques, to construct robust, 2-D scatter plots explaining why a model predicts a test example to be (1) in-distribution and of a particular class, (2) in-distribution but unsure of the class, and (3) out-of-distribution. ML consumers and engineers can visually compare the salient features of test samples with training examples through the use of a ``hover callback'' to understand model uncertainty performance and decide follow up courses of action.   We demonstrate the effectiveness of ScatterUQ to explain model uncertainty for a multiclass image classification on a distance-aware neural network trained on Fashion-MNIST and tested on Fashion-MNIST (in distribution) and MNIST digits (out of distribution), as well as a deep learning model for a cyber dataset. We quantitatively evaluate dimensionality reduction techniques to optimize our contextually driven UQ visualizations. Our results indicate that the ScatterUQ system should scale to arbitrary, multiclass datasets. 
Our code is available at \url{https://github.com/mit-ll-responsible-ai/equine-webapp}.
} 

\CCScatlist{
  Uncertainty quantification - Machine learning - Dimensionality reduction - Visualization - Explainable AI
}

\begin{document}


\firstsection{Introduction}

\maketitle

Deep neural network machine learning (ML) models are becoming increasingly popular for multiclass classification problems. However, the distribution of the data used to train ML models is often different than the distribution of real data that models encounter. When models encounter data ``in the wild'' that is confusing or dissimilar from the training data, it is important for models to be able to quantify their uncertainty to help end users make appropriate decisions and to maintain end user trust \cite{UQDL2022}. Otherwise, end users may blindly trust the model, make inappropriate decisions, or lose trust in the model altogether. Recently, \emph{distance-aware} neural networks have been developed that provide this extra uncertainty quantification (UQ) context alongside each model prediction, often in the form of a calibrated prediction probability and an out-of-distribution (OOD) score in addition to the class label \cite{OODreview,SNGP}. These models leverage distances between/among training and test data in the high-dimensional, penultimate neural network hidden layer to generate probabilities or OOD scores \cite{SNGP, Closeness2022, mahalanobis, quetal}. With dimensionality reduction (DR) and/or post-hoc clustering and fitting, the high-dimensional embeddings can lend themselves to helpful visualizations that generate intuition for the decision boundaries of multiple classes \cite{umap, Lamp, FeatMap, DeepLens, OODAnalyzer}. However, it is arguably not feasible or scalable to visualize all of the data in all usage scenarios, for arbitrary data (images, text, cyber, etc.), and for a large number of class labels (e.g., calibrated class probabilities may differ substantially from OOD scores).

In this paper, we present ScatterUQ, an interactive visualization system that explains model uncertainty by letting users compare test samples with relevant model training examples in specific uncertainty contexts. ScatterUQ is a visualization component in the open-source python module EQUINE\footnote{\url{https://github.com/mit-ll-responsible-ai/equine}}, which are together an ML system that makes UQ a primary consideration in model design and deployment.
ScatterUQ leverages the semantic distances in the high dimensional embeddings in a distance-aware neural network by using dimensionality reduction to visualize test samples and training examples as scatter/contour plots. The target end users of ScatterUQ include \textbf{ML Consumers} (including operators, analysts, and decision makers who decide downstream courses of action using ML models) and \textbf{ML Engineers}, who design, train, debug, and deploy ML models. 
Our main contributions include:
\begin{itemize}
    \item A proposed solution to the scalability problem of visualizing UQ information for an arbitrary number of classes, driven by interactive confidence sliders that filter test data into three UQ contexts: (1) high-confidence predictions on in-distribution data, comparing a test input to subsampled examples from its nearest class (2) low-confidence predictions on in-distribution data, comparing a test input to examples from a small number of classes, and  (3) out-of-distribution data, comparing a test example to its closest in-distribution class. 
    \item A comparison of dimensionality reduction techniques on high dimensional input embeddings to visualize one test sample at a time with a local, relevant contextual subset of training examples for optimal distance and neighborhood preservation. 
    \item An extensible, open source python library EQUINE, that adds uncertainty quantification to generic feedforward, multiclass neural networks, and ScatterUQ, a companion open-source ReactJS system that simplifies visualizations of new data.
\end{itemize}

We demonstrate that when reducing from a high dimensional embedding space to a 2-D visualization, using subsampled data for local plots that target specific UQ use cases (instead of global data views) improves quantitative metrics  for both distance and neighborhood preservation, as well as reducing computation time. We provide examples of the context-driven use cases in problems from the image domain (Fashion-MNIST \cite{Fashion-MNIST} and MNIST \cite{mnist}) as well as from a cyber domain. 
We envision end users leveraging the EQUINE open-source library to enhance their neural networks with uncertainty quantification and to use ScatterUQ to rapidly build downstream visualizations for ML Engineers and Consumers.

\section{Background}

A comprehensive review of visualization techniques for ML in 2021 identified opportunities in UQ visualizations to aid ML engineers during model refinement and to address concept drift, but did not incorporate  \cite{VizSurvey2021}. In parallel, as the ML community has identified and begun rectifying issues with UQ in deep, multiclass neural networks \cite{SNGP,Closeness2022,mahalanobis,quetal,OODreview,UQDL2022}, the visual analytics community has begun building techniques for specific problem domains that globally illuminate out-of-distribution data, class confidences, and decision boundaries, with \textsc{OoDAnalyzer} (image data) and \textsc{DeepLens} (text data) being most similar to our work \cite{OODAnalyzer, DeepLens, vizsec21poster}. In particular, \textsc{DeepLens} and our earlier work \cite{vizsec21poster} use a slider to control the OOD threshold, which provides an interactive threshold that is of particular importance for uncalibrated OOD scores \cite{UQDL2022}. Often, visualization goals extend beyond UQ considerations to include saliency maps or other explainable AI considerations that are beyond the scope of this work, since ScatterUQ is designed to be domain-agnostic. Additionally, ScatterUQ considers only visualizations for intra- and inter-class confidence, a kind of aleatoric/data uncertainty, and OOD scores, a type of epistemic uncertainty. It does not explicitly treat uncertainties arising from model weights, architectures, or adversarial ML.

ScatterUQ builds upon this prior work by tightly coupling UQ visualizations to training and test data representations in the embedding space of an uncertainty- and distance-aware neural network, most easily in conjunction with the EQUINE python module, which is designed to expeditiously outfit any feedforward, multiclass PyTorch model with UQ. ScatterUQ visualizations assume that the neural network produces meaningful embeddings with calibrated probabilities and OOD scores $\in [0,1]$. For this paper, we rely upon a Prototypical Neural Network (or \emph{protonet}) architecture and follow the approach of Jorgensen et al.\cite{quetal} to outfit it with an OOD score, although it is possible to support other distance-aware approaches such as the mean embeddings from a Spectral-normalized Neural Gaussian Process model \cite{SNGP}. 

Protonets are neural networks that classify examples based on their distance (in a learned embedding space) to class representations called \emph{prototypes}.  The prototype for a class is the average of several labeled training examples.  During training ``episodes'', we sample the training examples used to define the class prototypes and ``query'' examples over which we will compute the loss function for use in updating the network weights via stochastic gradient descent.  To be specific, the predicted probability of query example $x_i$ for class $k$ is
\begin{equation}
\label{eq:softmax}
\frac{\exp(-d(f(x_i),p_k))}{\sum_{k'}\exp(-d(f(x_i),p_{k'}))},
\end{equation}
where $p_k$ is the prototype for class $k$, $d$ is a distance function, and $f$ is the embedding function.

The distances in the embedding space are also used to detect OOD examples. We define outlier scores based on the distance to the closest prototype using the relative Mahalanobis distance \cite{mahalanobis}.  When this distance is large, the example is more likely to be OOD.  To ease interpretation, we approximate the distribution of these distances with a kernel density estimate using a portion of the training set.  At test time, we compare the distance of a new point to the closest prototype to this distribution to compute the probability that an in-distribution example is closer than the test point, which provides an interpretable OOD score that does not rely on a separate ``OOD dataset''.

\section{Design Goals}

Through interviews with industry ML engineers, we identified four design goals for ScatterUQ:

    \noindent
    \textbf{G1: User-configurable uncertainty tolerance} End users may have different uncertainty tolerances depending on their data and contexts \cite{UQDL2022}. We want to allow users to configure what model predictions are considered ``confident enough'' to be accepted.
    
    \noindent
    \textbf{G2: Compare training examples and test samples.} Protonets leverage semantic high dimensional embeddings to make predictions. Rather than inspecting model weights, we want to explain model predictions by comparing similarity or dissimilarity between an test sample and the model's training examples.
    
    \noindent
    \textbf{G3: Preserve neighborhood and distance semantics in 2-D space.} Reducing data points from a high dimensional space to 2-D can result in misleading plots. We wanted to be able to accurately preserve distances and neighborhoods between points so that points that are close/far in the high dimensional space are also close/far in the low 2-D space. We discuss our approach in \autoref{sec:local-plots}.

    \noindent
    \textbf{G4: Filter data points by use case.} Too many data points on screen can overwhelm users and conflict with G3. We want to focus on one test sample at a time and only pick a subset of relevant training points to show in context. We discuss our approach in \autoref{sec:use-cases}.
    

\section{Methodology}

\subsection{Model Output Data Structure}
We trained a protonet on Fashion-MNIST and tested the model on both MNIST and Fashion-MNIST test sets. The protonet also produces an adjustable number of randomly sampled training examples (10-20 in this paper, which we fix at test time). For each test sample and saved training example, the model outputs a calibrated class confidence score for each class, an OOD or outlier score in $[0,1]$, its neural network embedding vector, and values used to represent subject matter expert-chosen features that are depicted via mouse hover.
%
Below is an abbreviated example JSON output:
\begin{lstlisting}[language=json,numbers=none,basicstyle=\small\ttfamily]
{   "class_confidence_scores": {
        "Ankle Boot": 0.94, "Sneaker": 0.04, ... },
    "outlier_score": 0.24,
    "embeddings": [ 0.02, -0.28, 0.70, ... ],
    "img_src": "input image src string, if applicable"
    "json_src": "input data as JSON, if applicable"     }
\end{lstlisting}
The length of each of the \texttt{embeddings} vectors is the width of the last neural network layer. 
We use dimensionality reduction techniques to reduce these arbitrarily high-dimensional vectors to 2-D scatterplots that show the proximity of test samples and training examples.

\subsection{Dimensionality Reduction Tradeoffs}
\label{sec:local-plots}
Chari et al.\ discovered that popular DR techniques like t-SNE \cite{tsne} and UMAP \cite{umap} can be manipulated to output arbitrary shapes in 2-D space that misrepresent the high dimensional data, especially with large datasets \cite{genomics}. We hypothesized that rather than running a DR method on a large ``global'' dataset of points, we could avoid arbitrariness (G3) and better address end user needs (G4) by displaying local, targeted plots for each test sample using only a subset of relevant training examples. In this paper, we define a \emph{local plot} as a DR plot containing only 1 test sample with training examples from only 1 or 2 relevant classes at a time. We define a \emph{global plot} as a DR plot with many test samples and all the model's training examples. We quantitatively evaluate these plots by computing DR metrics used by Yang et al.\ \cite{FeatMap}, with local neighborhood-preserving measures being continuity and trustworthiness \cite{van2009dimensionality}, and global distance-preserving measures being normalized stress \cite{Lamp} and the goodness-of-fit of a Shepard diagram \cite{Espadoto2019TowardAQ}. We show that for a relevant dataset under three contextualized UQ scenarios, (1) Principal Component Analysis (PCA) outperforms other methods and (2) DR metrics under local usage scenarios tend to be better than in a global view. By explicitly limiting the number of concurrent classes and sampled training sizes (e.g., by treating an OOD datapoint as a binary decision to be included in its nearest class), these approaches will scale to arbitrary numbers of classes and training data.

To evaluate our hypothesis that local plots better represent the high dimensional data in 2-D space, we ran experiments computing four DR metrics in a variety of scenarios to compare the performance of PCA \cite{pca}, t-SNE, UMAP, and Multi-dimensional Scaling (MDS) \cite{mds}. We considered three local usage scenarios: U1 is a high-confidence test point, U2 is a low-confidence, in-distribution test-point, and U3 is an OOD test-point. A global view can theoretically handle all of these cases simultaneously, but we hypothesized that DR metric fidelity should be more robust in representing embedding manifolds from 1 or 2 classes instead of $N$ classes, even for $N=10$. To account for the effects of training example selection used for building the contours, we constructed 10 sets of randomly selected training examples, each containing 10 training examples and 1 prototype for every class in Fashion-MNIST. 

For the local plots, we ran samples from the MNIST and Fashion-MNIST test sets through each subset of training examples and selected 10 test samples from each test set and for each use case described in \autoref{sec:use-cases} (10 test samples $\times$ 3 use cases $\times$ 2 test sets $\times$ 10 sets of training samples = \textbf{600 total local plots}). For each test sample, we created a local DR plot with the training examples and prototypes from only one to two relevant classes at a time depending on the use case (1 test sample + (1 or 2 classes) $\times$ (10 training examples + 1 prototype per class) = \textbf{12 or 23 points per local plot}). 

We created \textbf{10 global plots}, one for each set of randomly chosen, representative training examples. We randomly selected 25 test samples from each class for both the MNIST and Fashion-MNIST test sets and added all the training examples and prototypes into one global plot (25 test samples $\times$ 10 classes $\times$ 2 test sets + 10 Fashion-MNIST classes $\times$ (10 training examples and 1 prototype per class) = \textbf{610 points per global plot}).

\subsection{Dimensionality Reduction Experiment Results}

For each method, we calculated the resulting embeddings' computation time, Continuity \cite{Venna2005LocalMS}, Normalized Stress \cite{Espadoto2019TowardAQ}, Shepard goodness of fit \cite{Espadoto2019TowardAQ}, and Trustworthiness \cite{Venna2005LocalMS} mean scores and standard deviations, shown in \autoref{tab:method_metrics} and \autoref{tab:pca_metrics}. \autoref{tab:method_metrics} demonstrates that for the local plots, PCA is much faster and better preserves distances and nearest neighbors across a wide variety of local visualizations. Based on these results, we chose to use PCA for all the local plots. \autoref{tab:pca_metrics} breaks out all three local usage scenarios to compare directly with a global visualization, when using PCA. In general, PCA performed similarly or substantially better in the local plots, which supports our hypothesis that when reducing from high to low dimensional data, using fewer data points can more accurately preserve distances and neighborhoods.

%

\begin{table}[tb]
    \centering
    \caption{Visualization metrics for multiple DR methods averaged over 3 local usage scenarios showing that PCA is most performant.}
    \label{tab:method_metrics}
    \scriptsize
    \resizebox{\linewidth}{!}{%
    \begin{tabu}{llllll}
\toprule
Method & 	Time (sec) $\downarrow$ & 	Stress $\downarrow$ & 	Shepard $\uparrow$& 	Continuity $\uparrow$ & 	Trust $\uparrow$	\\ \midrule
PCA   & 	\textbf{5.4e-4} ($\pm$ 6.4e-5) & 	\textbf{0.06} ($\pm$ 0.02) & 	\textbf{0.94} ($\pm$ 0.03) & 	\textbf{0.95} ($\pm$ 0.03) & 	\textbf{0.94} ($\pm$ 0.04)  	\\
t-SNE & 	1.5e-1 ($\pm$ 1.6e-2) & 	47.30 ($\pm$ 73.16) & 	0.82 ($\pm$ 0.06) & 	0.90 ($\pm$ 0.06) & 	0.89 ($\pm$ 0.06)  	\\
UMAP  & 	2.0e+0 ($\pm$ 4.2e-1) & 	0.60 ($\pm$ 0.17) & 	0.85 ($\pm$ 0.07) & 	0.89 ($\pm$ 0.06) & 	0.88 ($\pm$ 0.07)  	\\
MDS   & 	9.7e-3 ($\pm$ 1.6e-3) & 	0.91 ($\pm$ 0.02) & 	0.32 ($\pm$ 0.18) & 	0.60 ($\pm$ 0.08) & 	0.58 ($\pm$ 0.08)  	 \\
\bottomrule
\end{tabu}
    }
\end{table}

\begin{table}[tb]
    \centering
    \caption{Visualization metric comparisons across multiple usage scenarios using 2-D PCA as the DR technique; all local scenarios are better or on par with global visualizations.}
    \label{tab:pca_metrics}
    \scriptsize
    \resizebox{\linewidth}{!}{%
    \begin{tabular}{llllll}
\toprule
Domain & 	Time (sec) $\downarrow$ & 	Stress $\downarrow$ & 	Shepard $\uparrow$& 	Continuity $\uparrow$ & 	Trust $\uparrow$	\\ \midrule
Local, U1 & 	5.5e-4 ($\pm$ 3.1e-5) & 	0.05 ($\pm$ 0.02) & 	0.95 ($\pm$ 0.03) & 	0.97 ($\pm$ 0.01) & 	0.95 ($\pm$ 0.02)  	\\
Local, U2 & 	5.4e-4 ($\pm$ 1.0e-4) & 	0.07 ($\pm$ 0.02) & 	0.92 ($\pm$ 0.03) & 	0.94 ($\pm$ 0.01) & 	0.93 ($\pm$ 0.04)  	\\
Local, U3 & 	5.3e-4 ($\pm$ 1.0e-5) & 	0.06 ($\pm$ 0.02) & 	0.94 ($\pm$ 0.03) & 	0.94 ($\pm$ 0.03) & 	0.93 ($\pm$ 0.04)  	\\ \midrule
Global, all  & 	2.6e-3 ($\pm$ 1.2e-3) & 	0.14 ($\pm$ 0.005) & 	0.80 ($\pm$ 0.01) & 	0.96 ($\pm$ 0.002) & 	0.86 ($\pm$ 0.004)  \\
\bottomrule
\end{tabular}

    }
\end{table}

\section{User Interface Design}

When a protonet model finishes training, the model saves some training examples that can be used for ScatterUQ. After training, the user can upload test data to the model to make predictions.

\subsection{Confidence Sliders}

\begin{figure}[h]
    \centering
     \includegraphics[width=.48\textwidth]{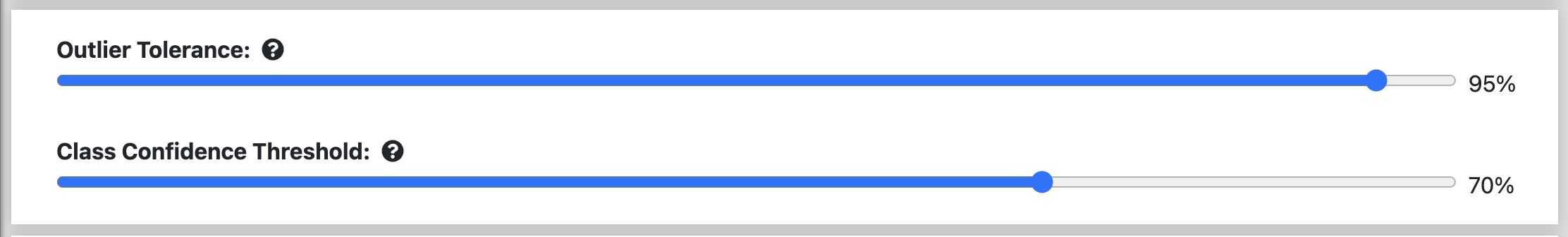}
    \caption{ScatterUQ confidence sliders that filter test samples into the High Confidence, Class Confusion, and OOD Use Cases}
    \label{fig:sliders}
\end{figure}

To support G1, ScatterUQ uses two confidence sliders shown in \autoref{fig:sliders} that determines to which use case a test sample belongs. The Outlier Tolerance slider filters out test samples that may be OOD. If the outlier score of a sample (e.g., 0.99) is greater than the user-defined outlier tolerance (e.g., 0.95), then the sample is reclassified as \texttt{OTHER} and filtered into the OOD Use Case (\autoref{sec:ood-use-case}). A high Outlier Tolerance produces more labeled predictions (possibly with more false positives); a low-tolerance setting filters out more predictions (possibly with more false negatives).

The second slider controls the Class Confidence Threshold. If the class confidence scores of a sample (e.g., Shirt: 0.5, T-shirt/top: 0.48, etc.) are all lower than the class confidence threshold (e.g., 0.7), then the sample is reclassified as \texttt{OTHER} and filtered into the Intra-Class Confusion Use Case (\autoref{sec:class-confusion-use-case}). Test samples that have a low enough outlier score and high enough class confidence score are filtered into the High Confidence Use Case described in \autoref{sec:high-confidence-use-case}.

\subsection{Local Plots}

To support G2, ScatterUQ displays a DR scatter plot for each test sample that allows users to compare the test sample with relevant training examples, seen in \autoref{fig:ood}. The left sidebar is fixed to the test sample and displays the original input, predicted class, class confidence scores, and outlier score. 

The scatter plot displays the test sample (large, colored or gray dot), several class training examples (small, colored dots) and prototypes (small black dots) after DR.  Confidence contours are calculated using the weighted density of the 2-D coordinates from a class's training examples, with weights being either ``1-outlier score'' for suspected OOD test samples (U3, below) or class confidences otherwise (U1 and U2 below). Confidence contours are calculated using D3 Contour\footnote{\url{https://github.com/d3/d3-contour}} and marching squares\footnote{\url{https://en.wikipedia.org/wiki/Marching\_squares}}. Each plot also displays DR metrics for advanced users: Continuity, Normalized Stress, the first five Scree (eigen)values, Shepard goodness of fit, and Trustworthiness.
When a user hovers over training examples, the right sidebar displays the training data, true class, class confidence scores, and outlier score. By hovering over different points, users can compare the test sample with training examples using the two sidebars.

\section{Use Cases}
\label{sec:use-cases}

ScatterUQ currently supports 3 major use cases depending on whether a test sample meets the minimum threshold set by the confidence sliders. For each use case, ScatterUQ filters the relevant training examples from the one or two closest classes, describes the model confidence or uncertainty, and suggests possible courses of action for ML consumers and engineers.

\subsection{U1: High-Confidence Use Case}
\label{sec:high-confidence-use-case}

\begin{figure}[h]
    \centering
     \includegraphics[clip, trim=0cm 0 0 0cm, width=.48\textwidth]{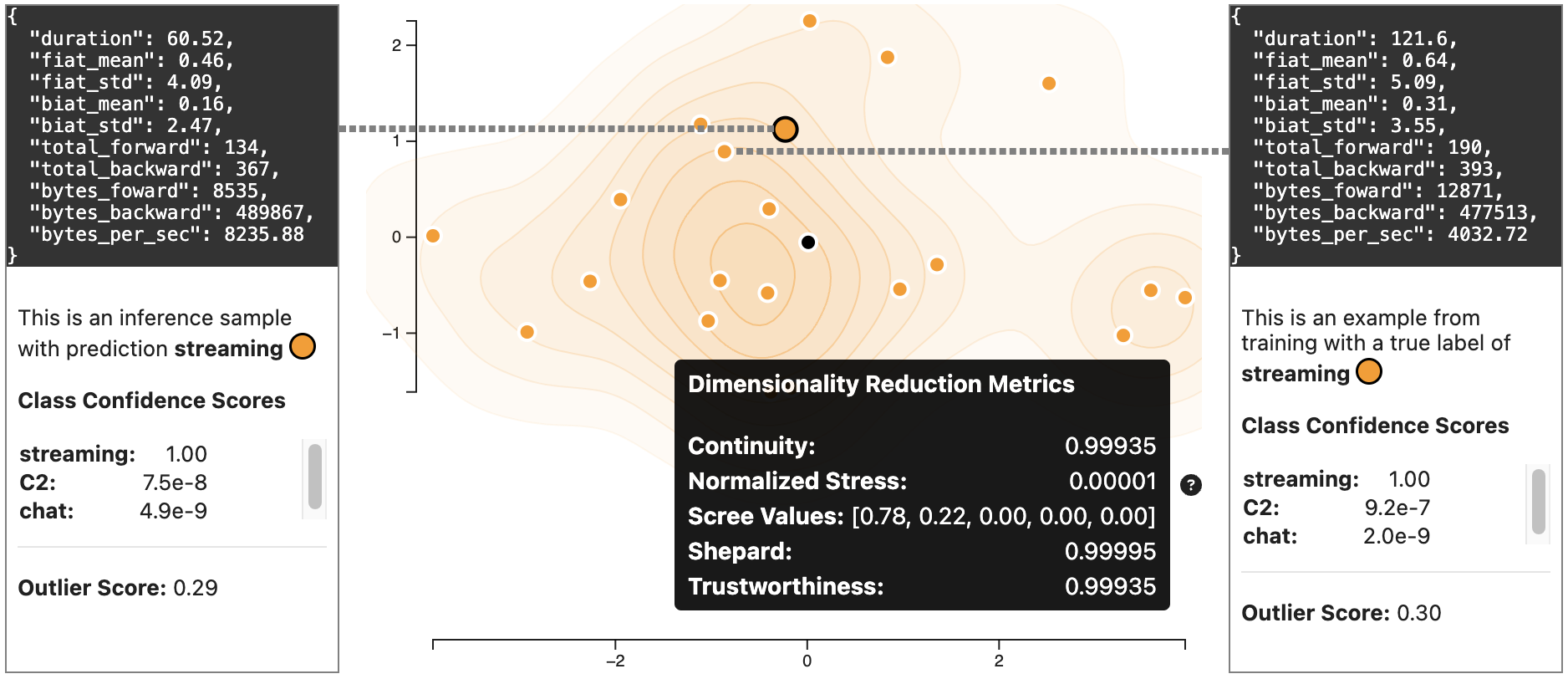}  
    \caption{High-confidence ScatterUQ plot of encrypted network traffic tabular features predicted to originate from a "streaming" application.}
    \label{fig:vnat}
\end{figure}

The first, simplest use case (U1) is when the model is confident enough that the sample is both in distribution and of a particular class. This visualization allows the ML Consumer to verify that the test point is in the neighborhood of training data, whose images can be observed by hovering over the training and test samples, and is a good sanity test for an ML Engineer. \autoref{fig:vnat} depicts an example of this scenario using a non-image dataset to both highlight the visualization (showing 20 training points, a test point, and the prototype) and the JSON text-driven hover data depicting cyber-relevant features for this encrypted traffic dataset \cite{quetal}.

\subsection{U2: Intra-Class Confusion Use Case}
\label{sec:class-confusion-use-case}

\begin{figure}[h]
    \centering
     \includegraphics[clip, trim=0cm 0 0 0cm, width=.48\textwidth]{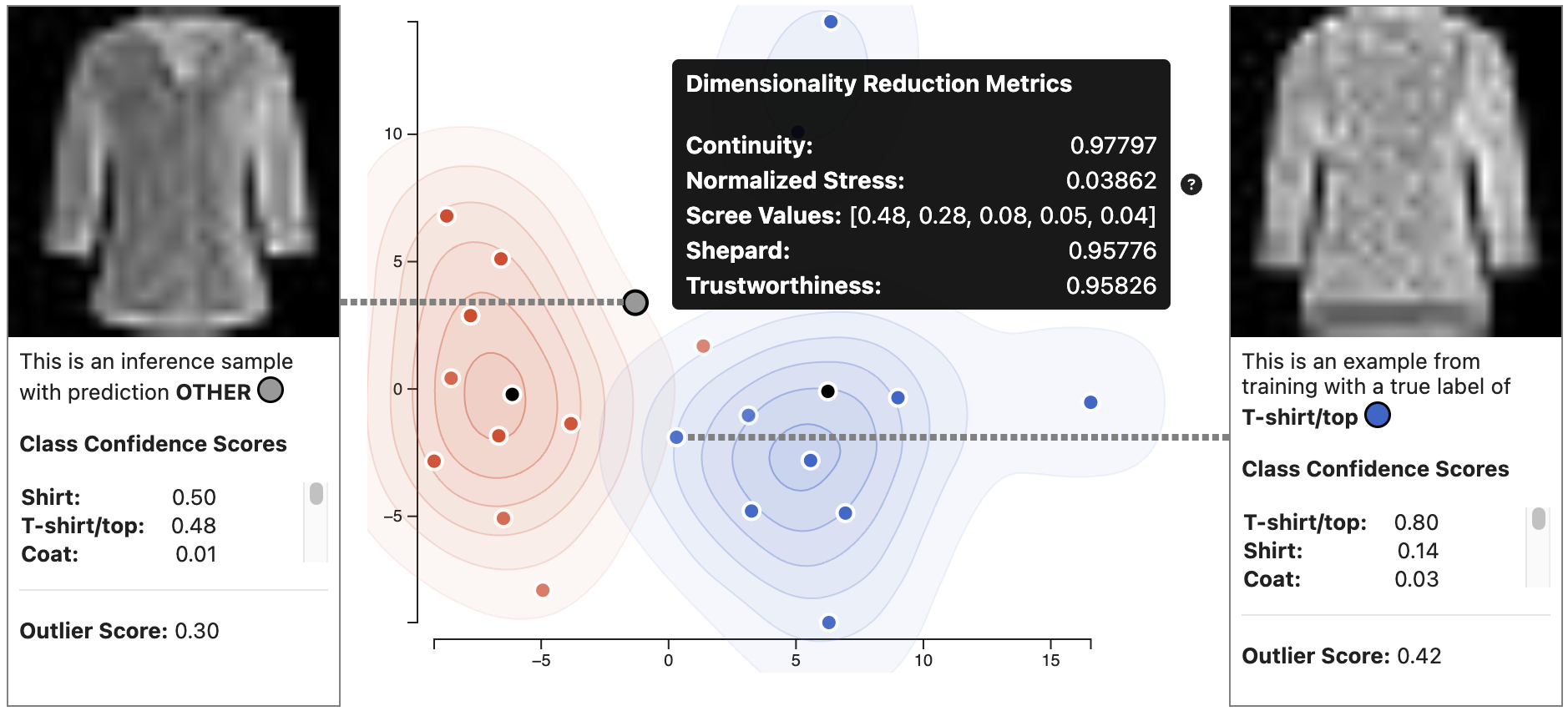}  
    \caption{ScatterUQ plot of a sample that is in-distribution but has low class confidence. ScatterUQ displays the test sample (left) and training examples of the two closest classes (\texttt{Shirt} in red and \texttt{T-shirt/top} in blue) as well as the closest training example (right).}
    \label{fig:confused}
\end{figure}

The second use case is when the model thinks the sample is in-distribution but is confused about which class the sample belongs to. In \autoref{fig:confused}, the model thinks the test sample is in distribution but is unsure whether the it belongs to the \texttt{Shirt} or \texttt{T-shirt/top} class. ScatterUQ displays a local PCA plot of the test sample and both the \texttt{Shirt} and \texttt{T-shirt/top} training examples, allowing the user to compare the test sample with the training examples from the two classes to make a final decision. In summary, the end user may ask: \textit{Why is the model confused about which class the sample belongs to?} ScatterUQ can help them conclude: The model is confused because the sample looks like data from both classes. ML Consumers know to be careful using the class predictions, and ML Engineers know they may need additional training data or to refactor their labels.

\subsection{U3: Out of Distribution (OOD) Use Case}
\label{sec:ood-use-case}

The third use case is when the model thinks that the sample is out of distribution. In \autoref{fig:ood}, the model correctly believes that the test sample (a digit 6 from the MNIST dataset) is OOD from its Fashion-MNIST training set. ScatterUQ displays a local PCA plot of the test sample and the training examples from the nearest class, \texttt{Sandals}. The user can compare the test sample with those training examples to determine if the sample is truly out of distribution. 
In our usage, we found that the model can consistently predict high outlier scores for MNIST samples and also assigns high class probabilities to the Sandal class. This makes sense because MNIST digits are genuinely OOD, but if we want the model to make its best guess, the thin curves and loops in the digits do most closely match with the thin loops in the Sandals class.
In summary, the end user may ask: \textit{Why does the model think this sample is out of distribution?} ScatterUQ can help them conclude: The model thinks the sample is OOD because it doesn't look like the training data. ML Consumers then know to be careful with the class predictions, and ML Engineers may be able to recognize the introduction of a new class label.

\section{Conclusions and Future Work}

We have presented ScatterUQ, a visualization system designed to complement distance- and uncertainty-aware, multiclass neural networks and provide context-driven visualizations for three different usage scenarios.  By focusing on local visualizations, ScatterUQ can scale to arbitrarily many classes and data sizes. Its confidence sliders allow it to fluidly adapt to a user's requirements. A quantitative comparison indicated that a simple dimensionality reduction technique (PCA) can quickly and robustly preserve the semantic distances in the embedding of a prototypical neural network. We highlighted three use cases --  high-confidence test points, low-confidence but in-distribution test-points, and low-confidence / out-of-distribution test points --  on datasets from two problem domains to demonstrate the generality of our approach. As future work, we intend to conduct evaluations of the efficacy of ScatterUQ for ML Engineers and Consumers on multiple, larger datasets.

\acknowledgments{
The authors wish to thank Gabriel Appleby and Remco Chang from Tufts University for their help with dimensionality reduction techniques.

DISTRIBUTION STATEMENT A. Approved for public release. Distribution is unlimited. This material is based upon work supported by the Department of the Air Force under Air Force Contract No. FA8702-15-D-0001. Any opinions, findings, conclusions or recommendations expressed in this material are those of the author(s) and do not necessarily reflect the views of the Department of the Air Force. (c) 2023 Massachusetts Institute of Technology. Delivered to the U.S. Government with Unlimited Rights, as defined in DFARS Part 252.227-7013 or 7014 (Feb 2014). Notwithstanding any copyright notice, U.S. Government rights in this work are defined by DFARS 252.227-7013 or DFARS 252.227-7014 as detailed above. Use of this work other than as specifically authorized by the U.S. Government may violate any copyrights that exist in this work.
}
\bibliographystyle{abbrv-doi}

\bibliography{main}
\end{document}